\documentclass{amia}
\usepackage{graphicx}
\usepackage[labelfont=bf]{caption}
\usepackage[superscript,nomove]{cite}
\usepackage{color}
\usepackage{multirow}
\usepackage{url} 
\begin{document}

\renewcommand{\thefootnote}{\roman{footnote}}

\title{Towards Automatic Bot Detection in Twitter for Health-related Tasks}

\author{Anahita Davoudi, PhD$^{1}$, Ari Z. Klein, PhD$^{1}$, Abeed Sarker, PhD$^{2}$, Graciela Gonzalez-Hernandez, PhD$^{1}$
}

\institutes{
    $^1$Department of Biostatistics, Epidemiology and 
Informatics, Perelman School of Medicine, 
University of Pennsylvania, Philadelphia, PA 19104\\
$^2$Department of Biomedical Informatics, School of Medicine,
Emory University, Atlanta, GA 30322\\
}

\maketitle

\noindent{\bf Abstract}\\
\textit{With the increasing use of social media data for health-related research, the credibility of the information from this source has been questioned as the posts may originate from automated accounts or "bots". While automatic bot detection approaches have been proposed, there are none that have been evaluated on users posting health-related information. In this paper, we extend an existing bot detection system and customize it for health-related research. Using a dataset of Twitter users, we first show that the system, which was designed for political bot detection, underperforms when applied to health-related Twitter users. We then incorporate additional features and a statistical machine learning classifier to significantly improve bot detection performance. Our approach obtains F$_1$-scores of $0.7$ for the "bot" class, representing improvements of $0.339$. Our approach is customizable and generalizable for bot detection in other health-related social media cohorts.
}


\section*{Introduction}
In recent years, social media has evolved into an important source of information for various types of health-related research. Social networks encapsulate large volumes of data associated with diverse health topics, generated by active user bases in continuous growth. Twitter, for example, has 330 million monthly active users worldwide 
that generate almost 500 million micro-blogs (tweets) per day.\footnote{\url{https://www.internetlivestats.com/twitter-statistics/}. Accessed: 05/06/2019.} For some years, the use of the platform to share personal health information has been growing, particularly amongst people living with one or more chronic conditions and those living with disability. Twenty percent of social network site users living with chronic conditions gather and share health information on the sites, compared with 12\% of social network site users who report no chronic conditions.\footnote{\url{https://www.pewinternet.org/2011/05/12/the-social-life-of-health-information-2011/}. Accessed: 08/15/2019.}
Social media data is thus being widely used for health-related research, for tasks such as adverse drug reaction detection \cite{sarker15rev}, syndromic surveillance \cite{chen2016}, subject recruitment for cancer trials \cite{reuter18}, and characterizing drug abuse \cite{sarker2016social}, to name a few. Twitter is particularly popular in research due to the availability of the public streaming API,\footnote{\url{https://developer.twitter.com/en/docs/tutorials/consuming-streaming-data.html}. Accessed: 05/06/2019.} which releases a sample of publicly posted data in real time. While early health-related research from social media focused almost exclusively on population-level studies, some very recent research tasks have focused on performing longitudinal data analysis at the user level, such as mining health-related information from cohorts of pregnant women \cite{Sarker}.

When conducting user-level studies from social media, one challenge is to ascertain the credibility of the information posted. Particularly, it is important to verify, when deriving statistical estimates from user cohorts, that the user accounts represent humans and not \textit{bots} (accounts that can be controlled to automatically produce content and interact with other profiles)\cite{ref8, varol}. Bots may spread false information by automatically \textit{retweeting} posts without a human verifying the facts or to influence public opinions on particular topics on purpose \cite{ref8, Bessi, Thomas}. For example, a recent study \cite{Bron} showed that the highest proportion of anti-vaccine content is generated by accounts with unknown or intermediate bot scores, meaning that the existing methods were not able to fully determine if they were indeed bots. Automatic bot detection techniques mostly rely on extracting features from users' profiles and their social networks \cite{Hu20142, wu}. Some studies have used Honeypot profiles on Twitter to identify and analyze bots \cite{ref17}, while other studies have analyzed social proximity \cite{ref10} or both social and content proximities \cite{Hu20142}, tweet timing intervals \cite{ref19}, or user-level content-based and graph-based features \cite{McCord}. However, in response to efforts towards keeping Twitter bot-free, bots have evolved and changed to overcome the detection techniques \cite{ref2}.  



The objectives of this study are to (i) evaluate an existing bot detection system on user-level datasets selected for their health-related content, and (ii) extend the bot detection system for effective application within the health realm. Bot detection approaches have been published in the past few years, but most of the code and data necessary for reproducing the published results were not made available \cite{Chavoshi, Gilani, Minnich}. The only system for which we found both operational code and data available, Botometer \cite{davis} (formerly BotOrNot), was chosen as the benchmark system for this study. To the best of our knowledge, this paper presents the first study on health-related bot detection. We have made the classification code and training set of annotated users available at (we will provide a URL with the camera-ready version of the paper).


\section{Methods}

\subsection{Corpus} 
To identify bots in health-related social media data, we retrieved a sample of $10,417$ users from a database containing more than $400$ million publicly available tweets posted by more than $100,000$ users who have announced their pregnancy on Twitter \cite{Sarker}. This sample is based on related work for detecting users who have mentioned various pregnancy outcomes in their tweets. Two professional annotators manually categorized the $10,417$ users as "bot," "non-bot," or "unavailable," based on their publicly available Twitter sites. Users were annotated broadly as "bot" if, in contrast to users annotated as "non-bot," they do not appear to be posting personal information. Users were annotated as "unavailable" if their Twitter sites could not be viewed at the time of annotation, due to modifying their privacy settings or being removed or suspended from Twitter. Based on $1000$ overlapping annotations, their inter-annotator agreement (IAA) was $\kappa$ = $0.93$ (Cohen’s kappa~\cite{Cohen}), considered "almost perfect agreement" \cite{viera}. Their IAA does not include disagreements resulting from the change of a user's status to or from "unavailable" in the time between the first and second annotations. Upon resolving the disagreements, $413$ $(4\%)$ users were annotated as "bot," $7849$ $(75.35\%)$ as "non-bot," and $20.69$ $(19.9\%)$ as "unavailable". 

\subsection {Classification}
We used the $8262$ "bot" and "non-bot" users in experiments to train and evaluate three classification systems. We split the users into $80\%$ (training) and $20\%$ (test) sets, stratified based on the distribution of "bot" and "non-bot" users. The training set includes $61,160,686$ tweets posted by $6610$ users, and the held-out test set includes $15,703,735$ tweets posted by $1652$ users. First, we evaluated Botometer on our held-out test set. Botometer is a publicly available bot detection system designed for political dot detection. It outputs a score between 0 and 1 for a user, representing the likelihood that a user is a bot. Second, we used the Botometer score for each user as a feature in training a gradient boosting classifier which is a decision tree-based ensemble machine learning algorithm with gradient boosting \cite{fri} and can be used to address class imbalance. To adapt the Botometer scores to our binary classification task, we set the threshold to $0.47$, based on performing 5-fold cross validation over the training set. To further address the class imbalance, we used the Synthetic Minority Over-sampling Technique (SMOTE)\cite{Chawla} to create artificial instances of "bot" users in the training set. We also performed 5-fold cross validation over the training set to optimize parameters for the classifier; we used \emph{exponential} as the loss function, set the number of estimators to $200$, and set the learning rate to $0.1$. Third, we used the classifier with an extended set of features that are not used by Botometer. Based on our manual annotation, we consider the following features to be potentially informative for distinguishing "bot" and "non-bot" users in health-related data:


\begin{itemize}
  \item \emph{Tweet Diversity}. Considering that "bot" users may re-post the same tweets, we used the ratio of a user's unique tweets to the total number of tweets posted by the user, where 0 indicates that the user has posted only the same tweet multiple times, and 1 indicates that each tweet is unique and has been posted only once. As Figure 1 illustrates, a subset of "bot" users (in the training set) have posted more of the same tweets than "non-bot" users.

 \item \emph{URL score}. During manual annotation, we found that "bot" users' tweets frequently contain URLs (\emph{e.g.}, advertisements for health-related products, such as medications), so we use the ratio of the number of a user's tweets containing a URL to the total number of tweets posted by the user.

  \item \emph{Mean Daily Posts}. Considering that "bot" users may post tweets more frequently than "non-bot" users, we measured the average and standard deviation of the number of tweets posted daily by a user. As Figure 1 illustrates, a subset of "bot" users post, on average, more tweets daily than "non-bot" users.

 \item \emph{Topics}. Considering that "bot" users may post tweets about a limited number of targeted topics, we used topic modeling to the measure the heterogeneity of topics in a user's tweets. We used Latent Dirichlet Allocation (LDA)\cite{Blei} to extract the top five topics from all of the users' $1000$ most recent tweets (or all the tweets if a user has posted less than $1000$ tweets), and used the mean of the weights of each topic across all of a user's tweets. 

 \item \emph{Mean Post Length}. Considering that the length of tweets may be different between "bot" and "non-bot" users, we used the mean word length and standard deviation of a user's tweets.

\begin{figure}[H]
  \centering
  \includegraphics[width=15cm,height=10cm,keepaspectratio,]{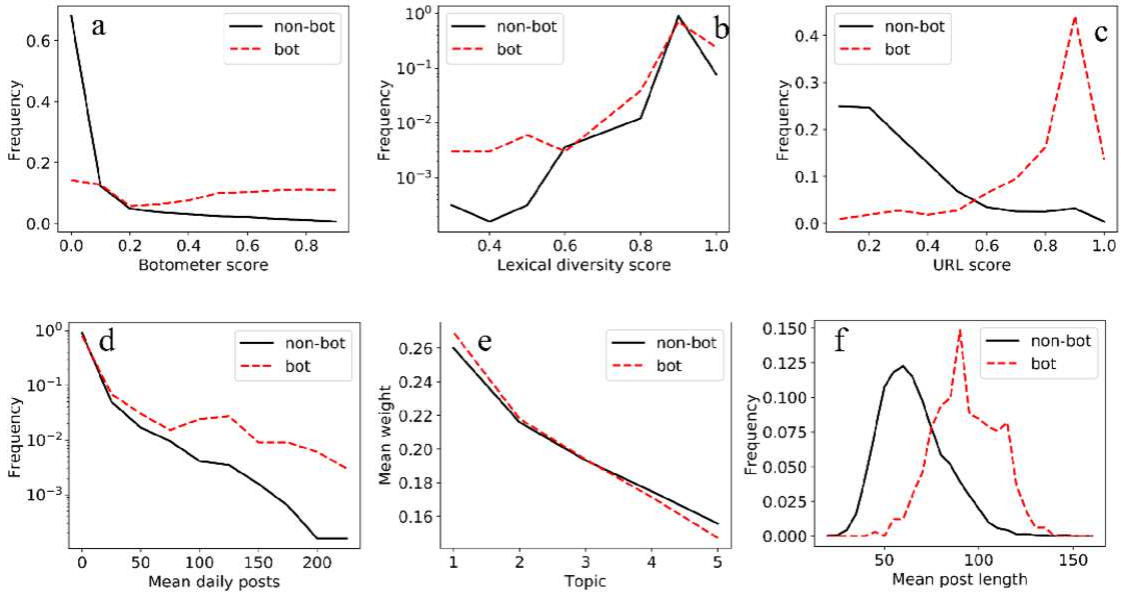}
  \caption{The distribution of features for "bot" and "non-bot" users in the training set. }
  \label{fig1}
\end{figure}
    
\item \emph{Profile Picture}. In addition to tweet-related features, we used features based on information in users' profiles. Considering that a "non-bot" user's profile picture may be more likely to contain a face, we used a publicly available system\footnote{\url{https://github.com/ageitgey/face_recognition}. Accessed: 08/15/2019.} to detect the number of faces in a profile picture. As Figure 2, illustrates a face was not detected in the profile picture of the majority of "non-bot" users (in the training set), whereas at least one face was detected in the profile picture of the majority of "bot" users.

\item \emph{User Name}. Finally, we used a publicly available lexicon\footnote{\url{https://pypi.org/project/gender-guesser/}. Accessed: 07/25/2019.} to detect the presence or absence of a person's name in a user name. As Figure 2 illustrates, the name of a person is present (1) in approximately half of "non-bot" user names, whereas the name of a person is absent (0) in the majority of "bot" user names.

\begin{figure}[H]
  \centering
  \includegraphics[width=15cm,height=10cm,keepaspectratio,]{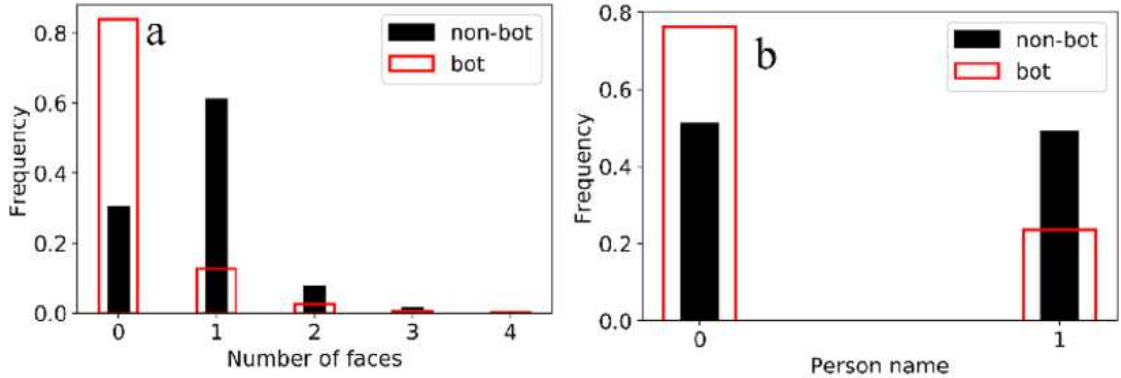}
  \caption{The distribution of faces detected in profile pictures and names detected in user names for "bot" and "non-bot" users in the training set.}
  \label{fig2}
\end{figure}

\end{itemize}


\section{Results}
Table 1 presents the precision, recall, and F$_1$-scores for the three bot detection systems evaluated on the held-out test set. The F$_1$-score for the "bot" class indicates that Botometer ($0.361$), designed for political bot detection, does not generalize well for detecting "bot" users in health-related data. Although the classifier with only the Botometer score as a feature ($0.286$) performs even worse than the default Botometer system, our extended feature set significantly improves performance ($0.700$). For imbalanced data, a higher F$_1$-score for the majority class is typical; in this case, it reflects that we have modeled the detection of "bot" users based on their natural distribution in health-related data.

\begin{table*}[!h]
\centering
\begin{tabular}{|c|c|c|c|c|c|}
 \hline
  Classifier & Precision (NB, B, Avg.) & Recall (NB, B, Avg.) & F$_1$-Score (NB, B, Avg.)\\
  \hline
  Botometer Default & 0.974, 0.276, 0.625 & 0.919, 0.542, 0.730 & 0.945, 0.361, 0.653\\
  \hline
  GB Classifier $_{Botometer score}$ & 0.963, 0.285, 0.624 & 0.962, 0.288, 0.625 & 0.962, 0.286, 0.624\\
  \hline
  GB Classifier $_{Botometer score+ Features}$ & 0.985, 0.678, 0.831 & 0.982, 0.724, 0.853 & 0.984, \textbf{0.700}, 0.842\\
  \hline

\end{tabular}
\label{table1}

\caption{Precision, recall, and F$_1$-score for three bot detection systems evaluated on a held-out test set of 1652 users. Precision, recall, and F$_1$-scores are reported for the "non-bot" class (NB), the "bot" class (B), and an average of the two classes (avg.).}
\end{table*}

\section{Discussion}
Our results demonstrate that (i) a publicly available bot detection system, designed for political bot detection, underperforms when applied to health-related data, and (ii) extending the system with simple features derived from health-related data significantly improves performance. An F$_1$-score of $0.700$ for the "bot" class represents a promising benchmark for automatic classification of highly imbalanced Twitter data and, in this case, for detecting users who are not reporting information about their own pregnancy on Twitter. Detecting such users is particularly important in the process of automatically selecting cohorts\cite{klein} from a population of social media users for user-level observational studies\cite{golder}.

A brief error analysis of the 25 false negatives users (in the held-out test set of 1652 users) from the classifier with the extended feature set reveals that, while only one of the users is an account that automatically re-posts other users' tweets, the majority of the errors can be attributed to our broad definition of "bot" users, which includes health-related companies, organizations, forums, clubs, and support groups that are not posting personal information. These users are particularly challenging to automatically identify as "bot" users because, with humans posting on behalf of an online maternity store, or to a pregnancy forum, for example, their tweets resemble those posted by "non-bot" users. In future work, we will focus on deriving features for modeling the nuances that distinguish such "bot" users.

\section{Conclusion}

As the use of social networks, such as Twitter, in health research is increasing, there is a growing need to validate the credibility of the data prior to making conclusions. The presence of bots in social media presents a crucial problem, particularly because bots may be customized to perpetuate specific biased or false information, or to execute advertising or marketing goals. We demonstrate that, while existing systems have been successful in detecting bots in other domains, they do not perform as well for detecting health-related bots. Using a machine learning algorithm on top of an existing bot detection system, and a set of simple derived features, we were able to significantly improve bot detection performance in health-related data. Introducing more features would likely contribute to further improving performance, which we will explore in future work.

\section*{Acknowledgments}
This study was funded in part by the National Library of Medicine (NLM) (grant number: R01LM011176) and the National Institute on Drug Abuse (NIDA) (grant number: R01DA046619) of the National Institutes of Health (NIH). The content is solely the responsibility of the authors and does not necessarily represent the official views of the National Institutes of Health.

\makeatletter
\renewcommand{\@biblabel}[1]{\hfill #1.}
\makeatother

\bibliographystyle{unsrt}

\end{document}